\documentclass[runningheads,a4paper]{article}

\usepackage{amssymb,amsmath}
\usepackage{graphicx}

\DeclareMathOperator*{\argmin}{arg\,min}

\usepackage{hyperref}

\newtheorem{thm}{Theorem}[section]
\newtheorem{lemma}[thm]{Lemma}

\newtheorem{definition}{Definition}[section]

\begin{document}


\title{From Preference-Based to Multiobjective \\Sequential Decision-Making}


%
%
\author{
Paul Weng\footnote{\href{mailto:paweng@cmu.edu}{paweng@cmu.edu},  \url{http://weng.fr}}
}
%


%
%

\maketitle

\begin{abstract}
In this paper, we present a link between preference-based and multiobjective sequential decision-making.
While transforming a multiobjective problem to a preference-based one is quite natural, the other direction is a bit less obvious.
We present how this transformation (from preference-based to multiobjective) can be done under the classic condition that preferences over histories can be represented by additively decomposable utilities and that the decision criterion to evaluate policies in a state is based on expectation.
This link yields a new source of multiobjective sequential decision-making problems (i.e., when reward values are unknown) and justifies the use of solving methods developed in one setting in the other one.

\end{abstract}

\section{Introduction}\label{sec:intro}

Reinforcement learning (RL) \cite{SuttonBarto98} has proved to be a powerful framework for solving sequential decision-making problems under uncertainty. For instance, RL has been used to build an expert backgammon player \cite{Tesauro95}, an acrobatic helicopter pilot \cite{AbbeelCoatesNg10},
a human-level video game player \cite{MnihKavukcuogluSilverRusuVenessBellemareGravesRiedmillerFidjelandOstrovskiPetersenBeattieSadikAntonoglouKingKumaranWierstraLeggHassabis15}.
RL is based on the Markov decision process model (MDP) \cite{Puterman94}.
In the standard setting, both MDP and RL rely on scalar numeric evaluations of actions (and thus histories and policies). 
However, in practice, those evaluations may not be scalar or may even not be available.

Often actions are rather valued on several generally conflicting dimensions.
For instance, in a navigation problem, these dimensions may represent duration, cost and length.
This observation has led to the extension of MDP and RL to multiobjective MDP (MOMDP) and RL (MORL) \cite{RoijersVamplewWhitesonDazeley13}.
In multiobjective optimization, it is possible to distinguish three interpretations for objectives.
The first one corresponds to single-agent decision-making problems where actions are evaluated on different criteria, like in the navigation example.
The second comes up when the effects of actions are uncertain and one then also wants to optimize objectives that correspond to probability of success or risk for instance.
The last interpretation is in multiagent settings where each objective represents the payoff received by a different agent. 
Of course, in one particular multiobjective problem, one may encounter objectives with different interpretations.

More generally, sometimes no numerical evaluation of actions is available at all.
In this case, inverse reinforcement learning (IRL) \cite{NgRussell00} has been proposed as an approach to learn a reward function from demonstration provided by a human expert who is assumed to use an optimal policy.
However, this assumption may be problematic in practice as humans are known not to act optimally.
A different approach, qualified as preference-based, takes as initial preferential information the comparisons of actions or histories instead of a reward function.
This direction has been explored in the MDP setting \cite{GilbertSpanjaardViappianiWeng15} and the RL setting where it is called preference-based RL (PBRL) \cite{AkrourSchoenauerSebag12,FurnkranzHullermeierChengPark12}.

This theoretic paper presents a short overview of some recent work on multiobjective and preference-based sequential decision-making with the goal of relating those two research strands.
The contribution of this paper is threefold.
We build a bridge between preference-based RL and multiobjective RL, 
and highlight new possible approaches for both settings.
In particular, our observation offers a new interpretation of an objective, which yields a new source of multiobjective problems.

The paper is organized as follows.
In Section~\ref{sec:background}, we recall the definition of standard MDP/RL, their extensions to the multiobjective setting and their generalizations to the preference-based setting.
In Section~\ref{sec:mopb}, we show how MORL can be viewed as a PBRL problem.
This then allows the methods developed for PBRL to be imported to the MORL setting.
Conversely, in Section~\ref{sec:pbmo}, we show how some structured PBRL can be viewed as an MORL, which then justifies the application of MORL techniques on those PBRL problems.
Finally, we conclude in Section~\ref{sec:conclusion}.

\section{Background and Related Work}\label{sec:background}

In this section, we recall the necessary definitions needed in the next sections while presenting a short review of related work.
We start with the reinforcement learning setting (Section~\ref{sec:rl}) and then present its extension to the multiobjective setting (Section~\ref{sec:morl}) and to the preference-based setting (Section~\ref{sec:pbrl}).

\subsection{Reinforcement Learning}\label{sec:rl}

A reinforcement learning problem is usually defined using the Markov Decision Process (MDP) model.
A {\em standard MDP}  \cite{Puterman94} is defined as a tuple $\langle S, A, T, R \rangle$ where:
\begin{itemize}
\item $S$ is a finite set of states,
\item $A$ is a finite set of actions,
\item $T : S \times A \times S \to [0, 1]$ is a transition function with $T(s, a, s')$ being the probability of reaching state $s'$ after executing action $a$ in state $s$,
\item $R : S \times A \to \mathbb R$ is a reward function with $R(s, a)$ being the immediate numerical environmental feedback received by the agent after performing action $a$ in state $s$.
\end{itemize}
In this framework, a {\em $t$-step history} $h_t$ is a sequence of state-action: 
\begin{align*}
h_t = (s_0, a_1, s_1, \ldots, s_t)
\end{align*}
where $\forall i=0, 1, \ldots, t, s_i \in S$ and $\forall i=1, 2, \ldots, t, a_i \in A$.
The value of such a history $h_t$ is defined as:
\begin{align*}
R(h_t) = \sum_{i=1}^t \gamma^{i-1} R(s_{i-1}, a_i)
\end{align*}
where $\gamma \in [0, 1)$ is a discount factor.
A {\em policy} specifies how to choose an action in every state.
A {\em deterministic} policy $\pi : S \to A$ is a function from the set of states to the set of actions, while a {\em randomized} policy $\pi : S \times A \to [0, 1]$  states the probability $\pi(s, a)$ of choosing an action $a$ in a state $s$. 

The {\em value function} of a policy $\pi$ in a state $s$ is defined as:
\begin{align}\label{eq:exp}
v^\pi(s) = \mathbb E\big[\sum_{t\ge 0} \gamma^t \mathcal R_t\big]
\end{align}
where $\mathcal R_t$ is a random variable defining the reward received at time $t$ under policy $\pi$ and starting in state $s$.
Equation~(\ref{eq:exp}) can be computed iteratively as the limit of the following sequence: $\forall s\in S$,
\begin{align} 
v^\pi_0(s) &= 0 \label{eq:bellman1}\\
v^\pi_{t+1}(s) &= R(s, \pi(s)) + \gamma \sum_{s' \in S} T(s, \pi(s), s') v^\pi_t(s')\enspace. \label{eq:bellman2}
\end{align}

In a standard MDP, an {\em optimal} policy can be obtained by solving the Bellman's optimality equations: $\forall s\in S,$
\begin{align} \label{eq:obellman}
v^\pi(s) &= \max_{a\in A} R(s, a) + \gamma \sum_{s' \in S} T(s, a, s') v^\pi(s')\enspace.
\end{align}
Many solution methods can be used \cite{Puterman94} to solve this problem exactly: for instance, value iteration, policy iteration, linear programming. Approaches based on approximating the value function for solving large-sized state space have also been proposed \cite{SuttonBarto98}.

Classically, in reinforcement learning (RL), it is assumed that the agent does not know the transition and reward functions.
In that case, an optimal policy has to be learned by interacting with the environment.
Two main approaches can be distinguished here \cite{SuttonBarto98}.
The first (called indirect or model-based method), tries to first estimate the transition and reward functions and then use an MDP solving method on the learned environment model (e.g., \cite{StrehlLittman09}).
The second (called direct or model-free method), searches for an optimal policy without trying to learn a model of the environment.

The preference model that describes how policies are compared in standard MDP/RL is defined as follows.
A history is valued by the discounted sum of rewards obtained along that history.
Then, as a policy in a state induces a probability distribution over histories, it also induces a probability distribution over discounted sums of rewards.
The decision criterion used to compare policies in standard MDP is then based on the expectation criterion.

Both MDP and RL assume that the environmental feedback from which the agent plans/learns a (near) optimal policy is a scalar numeric reward value.
In many settings, this assumption does not hold.
The value of an action may be determined over several often conflicting dimensions.
For instance, in the autonomous navigation problem, an action lasts a certain duration, has an energy consumption cost and travels a certain length.
To tackle those situations, MDP and RL have been extended to deal with vectorial rewards.

\subsection{Multiobjective RL}\label{sec:morl}

Multiobjective MDP (MOMDP) \cite{RoijersVamplewWhitesonDazeley13} is an MDP $\langle S, A, T, \vec R \rangle$ where the reward function is redefined as $\vec R : S \times A \to \mathbb R^d$ with $d$ being the number of objectives.
The value function $\vec{v}^{\,\pi}$ of a policy $\pi$ is now vectorial and can be computed as the limit of the vectorial version of (\ref{eq:bellman1}) and (\ref{eq:bellman2}): $\forall s\in S$,
\begin{align} 
\vec{v}^{\,\pi}_0(s) &= (0, \dots, 0) \in \mathbb R^d \label{eq:vbellman1}\\
\vec{v}^{\,\pi}_{t+1}(s) &= \vec R(s, \pi(s)) + \gamma \sum_{s' \in S} T(s, \pi(s), s') \vec{v}^{\,\pi}_t(s')\enspace. \label{eq:vbellman2}
\end{align}
In MOMDP, the value function of policy $\pi$ {\em Pareto-dominates} that of another policy $\pi'$ if in every state $s$, $\vec{v}^{\,\pi}(s)$ is not smaller than $\vec{v}^{\,\pi'}(s)$ on every objective and $\vec{v}^{\,\pi}(s)$ is greater than $\vec{v}^{\,\pi'}(s)$ on at least one objective.
By extension, we say that $\pi$ Pareto-dominates $\pi'$ if value function $v^\pi$ Pareto-dominates value function $v^{\pi'}$.
A value function (resp. policy) is {\em Pareto-optimal} if it is not Pareto-dominated by any other value function (resp. policy).
Due to incomparability of vectorial value functions, there are generally many Pareto-optimal value functions (and therefore policies), which constitutes the main difficulty of the multiobjective setting.

Similarly to standard MDP, MOMDP can be extended to multiobjective reinforcement learning (MORL), in which case the agent is not assumed to know the transition function, neither the vectorial reward function.

In multiobjective optimization, four main families of approaches can be distinguished.
One first natural approach is to determine the set of all Pareto-optimal solutions (e.g., \cite{White82,LizotteBowlingMurphy10}). 
However, in practice, searching for all the Pareto-optimal solutions may not be feasible.
Indeed, it is known \cite{PernyWengGoldsmithHanna13UAI} that this set can be exponential in the size of the state and action spaces.
A more practical approach is then to determine an $\epsilon$-cover of it \cite{ChatterjeeMajumdarHenzinger06,PernyWengGoldsmithHanna13UAI}, which is an approximation of the set of Pareto-optimal solutions.
\begin{definition}
A set $C \subseteq \mathbb R^d$ is an {\em $\epsilon$-cover} of a set $P \subseteq \mathbb R^d$ if 
\begin{align*}
	\forall v \in P, \exists v' \in C, (1+\epsilon) v' \ge v 
\end{align*}
where $\epsilon>0$.
\end{definition}

Another approach related to the first one is to consider refinements of Pareto dominance, such as Lorenz dominance (which models a certain notion of fairness) or lexicographic order \cite{WrayZilbersteinMouaddib15,GaborKalmarSzepesvari98}.
In fact, with Lorenz dominance, the set of optimal value functions may still be exponential in the size of the state and action spaces.
Again, one may therefore prefer to determine its $\epsilon$-cover \cite{PernyWengGoldsmithHanna13UAI} in practice.

Still another approach to solve multiobjective problems is to assume the existence of a scalarizing function $f : \mathbb R^d \to \mathbb R$, which, given a vector $v \in \mathbb R^d$, returns a scalar valuation.
Two cases can be considered: $f$ can be either linear \cite{BarrettNarayanan08} or nonlinear \cite{PernyWeng10,OgryczakPernyWeng13,OgryczakPernyWeng11ADT}.

The scalarizing function can be used at three different levels:
\begin{itemize}
\item It can be directly applied on the vectorial reward function leading to the definition of a scalarized reward function.
This boils down to defining a standard MDP/RL from a MOMDP/MORL, which can then be tackled with standard solving methods.
\item It can also aggregate the different objectives of the vector values of histories and then a policy in a state can be valued by taking the expectation of those scalarized evaluation of histories.
\item It can be applied on the vectorial value functions of policies in order to obtain scalar value functions.
\end{itemize}
For linear scalarizing functions, those three levels lead to the same solutions.
However, for nonlinear scalarizing functions, they generally lead to different solutions.
In practice, it generally only makes sense to use a nonlinear scalarizing function on expected discounted sum of vector rewards (i.e., vector value functions), as the scalarizing function is normally defined to aggregate over the final vector values.
To the best of our knowledge, most previous work has applied a scalarizing function in this fashion.
In Section~\ref{sec:mopb}, we describe a setting where applying a nonlinear scalarizing function on vector values of histories could be justified.

A final approach to multiobjective problem assumes an interactive setting where a human expert is present and can provide additional preferential information (i.e., how to trade-off between different objectives).
This approach loops between the following two steps until a certain criterion is satisfied (e.g., the expert is satisfied with a proposed solution or there is only one solution left):
\begin{itemize}
\item show potential solutions or ask query to the expert
\item receive a feedback/answer from the expert
\end{itemize}
The feedback/answer from the expert allows to guide the search for a preferred solution among all Pareto-optimal ones \cite{SteuerChoo83}, or elicit unknown parameters of user preference model \cite{ReganBoutilier11e}. 


In both standard MDP/RL and MOMDP/MORL, it is assumed that numeric environmental feedback is available.
In fact, this may not be the case in some situations.
For instance, in the medical domain, it may be difficult and even impossible to value a treatment of a life-threatening illness in terms of patient well-being or death with a single numeric value.
Preference-based approaches have been proposed to handle these situations.

\subsection{Preference-Based RL}\label{sec:pbrl}

A preference-based MDP (PBMDP) is an MDP where possibly no reward function is given. 
Instead, one assumes that a preference relation $\gtrsim$ is defined over histories.
In the case where the dynamics of the system is not known, this setting is referred to as preference-based reinforcement learning (PBRL) \cite{FurnkranzHullermeierChengPark12,AkrourSchoenauerSebag12,BusaFeketeSzorenyiWengChengHullermeier13EWRL,BusaFeketeSzorenyiWengChengHullermeier14}.
Due to this ordinal preferential information, it is not possible to directly use the same decision criterion based on expectation like in the standard or multiobjective cases.
Most approaches in PBRL \cite{FurnkranzHullermeierChengPark12,BusaFeketeSzorenyiWengChengHullermeier13EWRL,BusaFeketeSzorenyiWengChengHullermeier14} relies on comparing policies with {\em probabilistic dominance}, which is defined as follows:
\begin{align}\label{eq:pd}
\pi \succsim \pi' \iff \mathbf P[\pi \gtrsim \pi'] \ge \mathbf P[\pi' \gtrsim \pi]
\end{align}
where $\mathbf P[\pi \gtrsim \pi']$ denotes the probability that policy $\pi$ generates a history preferred or equivalent to that generated by policy $\pi'$.
Probabilistic dominance is related to Condorcet methods (where a candidate is preferred to another if more voters prefers the former than the latter) in social choice theory.
This is why the optimal policy for probabilistic dominance is often called a {\em Condorcet winner}.

The difficulty with this decision model is that it may lead to preference cycles (i.e., $\pi \succ \pi' \succ \pi'' \succ \pi$) \cite{GilbertSpanjaardViappianiWeng15}.
To tackle this issue, three approaches have been considered.
The first approach simply consists in assuming some consistency conditions that forbid the occurence of preference cycles.
This is the case in the seminal paper \cite{YueBroderKleinbergJoachims12} that proposed the framework of dueling bandits.
This setting is the preference-based version of multi-armed bandit, which is itself a special case of reinforcement learning.
The second approach consists in considering stronger versions of (\ref{eq:pd}).
Drawing from voting rules studied in social choice theory, refinements such as Copeland's rule or Borda's rule for instance, have been considered \cite{BusaFeketeSzorenyiWengChengHullermeier13ICML,BusaFeketeSzorenyiWengChengHullermeier14}.
The last approach, which was proposed recently \cite{GilbertSpanjaardViappianiWeng15,DudikHofmannSchapireSlivkinsZoghi15}, consists in searching for an optimal mixed\footnote{The randomization is over policies and not over actions, like in randomized policies.} policy instead of an optimal deterministic policy, which may not exist.
Drawing from the minimax theorem in game theory, it can be shown that an optimal mixed policy is guaranteed to exist.

\section{MORL AS PBRL}\label{sec:mopb}

An MOMDP/MORL problem can obviously be seen as a PBMDP/PBRL problem.
Indeed, the preference relation $\gtrsim$ over histories can simply be taken as the preference relation induced over histories by Pareto dominance.
Then probabilistic dominance (\ref{eq:pd}) in this setting can be interpreted as follows.
A policy $\pi$ is preferred to another policy $\pi'$ if the probability that $\pi$ generates a history that Pareto-dominates a history generated by $\pi'$ is higher than the probability of the opposite event.
A minor issue in this formulation is that incomparability is treated in the same way as equivalence.

More interestingly, when a scalarizing function $f$ is given, scalarized values of histories can then be used and compared in (\ref{eq:pd}), leading to:
\begin{align*}
\pi \succsim \pi' \iff \mathbf P[f(\vec R(H_\pi)) \ge f(\vec R(H_{\pi'}))] \ge \mathbf P[f(\vec R(H_{\pi'})) \ge f(\vec R(H_\pi))]
\end{align*}
where $H_\pi$ (resp. $H_{\pi'}$) is a random history generated by policy $\pi$ (resp. $\pi'$) and $\vec R(H_\pi)$ (resp. $\vec R(H_{\pi'})$) is its vectorial value.
Notably, this setting motivates the application of a nonlinear scalarizing function on vector values of histories, which has not been investigated before \cite{RoijersVamplewWhitesonDazeley13}.

More generally, viewing MOMDP/MORL as a PBMDP/PBRL, one can import all the techniques and solving methods that have been developed in the preference-based settings \cite{FurnkranzHullermeierChengPark12,BusaFeketeSzorenyiWengChengHullermeier14,GilbertSpanjaardViappianiWeng15}.
As far as we know, both cases above (with Pareto dominance or with a scalarizing function) have not been investigated. 
We expect that efficient solving algorithms exploiting the additively decomposable vector rewards could possibly be designed by adapting PBMDP/PBRL algorithms.

When transforming a multiobjective into a preference-based problem, the decision criterion has generally to be changed from one based on expectation to one based on probabilistic dominance.
This change may be justified for different reasons.
For instance, when it is known in advance that an agent is going to face the decision problems only a limited number times, the expectation criterion may not be suitable because it does not take into account notions of variability and risk attitudes.
Besides, when the decision problem really corresponds to a competitive setting, 
probabilistic dominance is particularly well-suited.


\section{PBRL AS MORL}\label{sec:pbmo}

While viewing MOMDP/MORL as PBMDP/PBRL is quite natural, the other way around may be less obvious and more interesting.
We therefore develop in more details this direction by focusing on one particular case of PBMDP/PBRL where the preference relation over histories is assumed to be representable by an additively decomposable utility function and the decision criterion is based on expectation (e.g., as assumed in inverse reinforcement learning \cite{NgRussell00}).
This amounts to assuming the existence of a reward function $\hat R : S \times A \to \{x_1, \ldots, x_d\}$ where the $x_i$'s are unknown scalar numeric reward values.
Exploiting this assumption, we present two cases where PBMDP/PBRL can be transformed into MOMDP/MORL, and justifies the use of one scalarizing function, the Chebyshev norm, on the MOMDP/MORL model obtained from a PBMDP/PBRL model.

\subsection{From Unknown Rewards to Vectorial Rewards}\label{sec:relation}

The first transformation assumes that an order over unknown rewards is known, while the second assumes more generally that an order over some histories are known.

\subsubsection{Ordered Rewards}

In the first case, we assume that we know the order over the $x_i$'s.
Without loss of generality, we assume that $x_1 < x_2 < \ldots < x_d$.

Following previous work \cite{Weng11l,Weng12l}, it is possible to transform a PBMDP into an MDP with vector rewards by defining the following vectorial reward function $\bar R$ from $\hat R$:
\begin{align}\label{eq:breward}
\forall s\in S, \forall a\in A, \bar R(s, a) = \mathbf 1_i \mbox{ if } \hat R(s, a) = x_i
\end{align}
where $\mathbf 1_i$ is the $i$-th canonical vector of $\mathbb R^d$.
Using $\bar R$, one can compute the vector value function of a policy by adapting (\ref{eq:vbellman1}) and (\ref{eq:vbellman2}).
The $i$-th component of a vector value function of a policy $\pi$ in a state can be interpreted as the expected discounted count of reward $x_i$ obtained when applying policy $\pi$.
However, note that because of the preferential order over components, two vectors cannot be directly compared with Pareto dominance.
Another transformation is needed to obtain a usual MOMDP.

Given a vector $v$, we define its decumulative $v^\downarrow$ as follows:
\begin{align*}
\forall k=1, \ldots, d, v^\downarrow_k = \sum_{j=k}^d v_j
\end{align*}
A PBMDP/PBRL can be reformulated as the following MOMDP/MORL where the reward function is defined by:
\begin{align}\label{eq:vreward}
\forall s\in S, \forall a\in A, \vec R(s, a) = \mathbf{1}^\downarrow_i \mbox{ if } \hat R(s, a) = x_i
\end{align}
Using this reward function, the vector value function $\vec{v}^{\,\pi}$ of a policy $\pi$ can be computed by adapting (\ref{eq:vbellman1}) and (\ref{eq:vbellman2}).
One may notice that $\vec{v}^{\,\pi}(s)$ is the decumulative vector computed from $\bar{v}^{\pi}$.

The relations between the standard value function $v^\pi$, the vectorial value functions $\bar v^\pi$ and $\vec{v}^{\,\pi}$ are stated in the following lemma.
\begin{lemma}
We have:
\[\forall s\in S, v^\pi(s) = (x_1, x_2, \ldots, x_d) \cdot \bar v^\pi(s) = (x_1, x_2 - x_1, \ldots, x_d - x_{d-1}) \cdot \vec{v}^{\,\pi}(s)\]
where $x \cdot y$ denotes the inner product of vector $x$ and vector $y$.
\end{lemma}
It is then easy to see that if $\vec{v}^{\,\pi}(s)$ Pareto-dominates $\vec{v}^{\,\pi'}(s)$ then $v^\pi(s) \ge v^{\pi'}(s)$ thanks to the order over the $x_i$'s.

\subsubsection{Ordered Histories}

In some situations, the order over unknown rewards may not be known and may not be easily determined.
For instance, in a navigation problem, it may not be obvious how to compare each action locally.
However, comparing trajectories may be more natural and easier to perform for the system designer.
Note that although vectorial reward function $\bar R$ in (\ref{eq:breward}) can be defined, without the order over rewards $x_i$'s, vectorial reward function $\vec R$ in (\ref{eq:vreward}) (and thus the corresponding MOMDP/MORL) cannot be defined anymore.

In those cases, if sufficient preferential information over histories is given, the previous trick can be adapted using simple linear algebra.
We now present this new transformation from PBMDP/PBRL to MOMDP/MORL.
We assume that the following comparisons are available:
\begin{align}
h_1 \prec h_2 \prec \ldots \prec h_d
\end{align}
where the $h_i$'s are histories.
Using the vector reward $\bar R$, one can compute the vector value of each history, i.e., 
$\forall i=1, 2, \ldots, d$, if $h_i = (s_0, a_1, s_1, \ldots, s_t)$ then its value is defined by:
\begin{align*} 
\bar r_i = \sum_{j=1}^t \gamma^{j-1} \bar R(s_{j-1}, a_j) \in \mathbb R^d.
\end{align*}
We assume that $\{\bar r_1, \ldots, \bar r_d\}$ form an independent set, which implies that the matrix $H$ whose columns are composed of $\{\bar r_1, \ldots, \bar r_d\}$ is invertible.
Recall $H$ is the basis change matrix from basis $\{\bar r_1, \ldots, \bar r_d\}$ to the canonical basis $\{\mathbf 1_1, \ldots, \mathbf 1_d\}$ and its inverse matrix $H^{-1}$ is the basis change matrix in the other direction.
Rewards $x_i$'s represented by the canonical basis can then be expressed in the basis formed by the independent vectors $\{\bar r_1, \ldots, \bar r_d\}$ using the basis change matrix $H^{-1}$.
Now, let us define a new vector reward function $\vec R_H$ by:
\begin{align}
\forall s\in S, \forall a\in A, \vec R_H(s, a) = H^{-1\downarrow}_i \mbox{ if } \hat R(s, a)=x_i
\end{align}
where $H^{-1\downarrow}_i$ is the decumulative of the $i$-th column of matrix $H^{-1}$.
Using this new reward function, one can define vector value function $\vec{v}^{\,\pi}$ of a policy $\pi$ by adapting (\ref{eq:vbellman1}) and (\ref{eq:vbellman2}).
\begin{lemma}
We have:
\[\forall s\in S, v^\pi(s) = (r_1, r_2 - r_1, \ldots, r_d - r_{d-1}) \cdot \vec{v}^{\,\pi}(s)\]
where $r_i$ is the value of history $h_i$, i.e., $r_i = (x_1, \ldots, x_d) \cdot \bar r_i$.
\end{lemma}
As the value of the $r_i$'s is increasing with $i$, if $\vec{v}^{\,\pi}(s)$ Pareto-dominates $\vec{v}^{\,\pi'}(s)$, then $\pi$ should be preferred.

\subsubsection{Applying MORL techniques to PBRL}

We have seen two cases where a PBMDP/PBRL problem can be transformed into an MOMDP/MORL problem.
As a side note, one may notice that the second case is a generalization of the first one.
Thanks to this transformation, the multiobjective approaches that we recalled in Section~\ref{sec:morl} can be applied in the preference-based setting.
We now mention a few cases that would be interesting to investigate in our opinion.

Here, a Pareto-optimal solution corresponds to a policy that is optimal for admissible reward values that respects the order known over rewards or histories.
Like in MOMDP/MORL, it may not be feasible to determine the set of all Pareto optimal solutions.
A natural approach \cite{PernyWengGoldsmithHanna13UAI} is then to compute its $\epsilon$-cover to obtain a representative set of solutions that are approximately optimal.

Another approach is to use a non-linear scalarizing function like the Chebyshev distance to an ideal point.
A policy $\pi^*$ is {\em Chebyshev-optimal} if it minimizes:
\begin{align}\label{eq:chebyshev}
\pi^* = \argmin_\pi \max_{i=1,\ldots, d} \mathbf I_i - \sum_{s \in S} \mu(s) \vec{v}^{\, \pi}_i(s)
\end{align}
where $\mathbf I_i = \max_\pi \sum_{s \in S} \mu(s) \vec v^{\, \pi}_i(s)$ defines the $i$-th component of the ideal point $\mathbf I \in \mathbb R^d$, $\mu$ is a positive probability distribution over initial states and $\vec v^{\, \pi}_i$ is the $i$-th component of the vector value function of an MOMDP/MORL obtained from a PBMDP/PBRL.
It is possible to show that a Chebyshev-optimal policy is a {\em minimax-regret-optimal} policy \cite{ReganBoutilier11}, whose definition can be written as follows:
\begin{align}\label{eq:mmr}
\pi^* = \argmin_\pi \max_{x \in \mathcal R} \max_{\pi'}  \sum_{s \in S}  \mu(s) x \cdot \vec v^{\, \pi'}(s) - \sum_{s \in S} \mu(s) x \cdot \vec v^{\, \pi}(s)
\end{align}
where $\mathcal R \subset [0, 1]^d$ is the set of nonnegative values representing differences of consecutive reward values.
\begin{lemma}
A policy is Chebyshev-optimal if and only if it is minimax-regret-optimal.
\end{lemma}
It is easy to see that the maximum (over $x$) in (\ref{eq:mmr}) is attained by choosing $x$ as a canonical vector and equal to the maximum (over $i$) in (\ref{eq:chebyshev}).
This simple property justifies the application of one simple non-linear scalarizing function used in multiobjective optimization in the preference-based setting.

The interactive approach mentioned in Section~\ref{sec:morl} has been already exploited for eliciting the unknown rewards in interactive settings where comparison queries can be issued to an expert by interleaving optimization/learning phases with elicitation phases in PBMDP with value iteration \cite{WengZanuttini13,GilbertSpanjaardViappianiWeng15ADT} and PBRL with Q-learning \cite{WengBusaFeketeHullermeier13}. 
It would be interesting to use an interactive approach to elicit the reward values by comparing the element of an $\epsilon$-cover of the Pareto optimal solutions.
This technique may help reduce the number of queries.

\section{Conclusion}\label{sec:conclusion}

In this paper, we highlighted the relation between two sequential decision-making settings: preference-based MDP/RL and multiobjective MDP/RL.
In particular, we showed that multiobjective problems can also arise in situations of unknown reward values.
Based on the link between both formalisms, one can possibly import techniques designed for one setting to solve the other.
To illustrate our points, we also listed a few interesting cases.

Besides, in our translation of a PBMDP/PBRL to an MOMDP/MORL, we assumed that rewards were Markovian, which may not always be true in practice.
It would be interesting to extend our translation to the non-Markovian case \cite{GrettonPriceThiebaux03}.



\bibliographystyle{splncs03} 
\bibliography{biblio160226}

\end{document}